\documentclass[letterpaper]{article} 
\usepackage{aaai23}  
\usepackage{times}  
\usepackage{helvet}  
\usepackage{courier}  
\usepackage[hyphens]{url}  
\usepackage{graphicx} 
\usepackage{natbib}  
\usepackage{caption} 
\usepackage{algorithm}
\usepackage{algorithmic}
\usepackage{multirow}
\usepackage{amsmath}
\usepackage{makecell}
\usepackage{times}
\usepackage{helvet}
\usepackage{courier}
\usepackage{booktabs}
\usepackage{multirow}
\usepackage{graphicx}
\usepackage{array}
\usepackage{amssymb}
\usepackage{utfsym}
\usepackage{bbding}
\usepackage{pifont}
\usepackage{wasysym}
\usepackage{utfsym}
\usepackage{fontawesome}
\usepackage{makecell}
\usepackage{mathrsfs}
\usepackage{amsmath}
\usepackage{bm}
\usepackage[cmintegrals]{newtxmath}
\usepackage{caption}
\usepackage{graphicx}
 
\usepackage{subcaption}
\usepackage{colortbl}
\definecolor{mygray}{gray}{.9}
\usepackage{arydshln}
\usepackage{newfloat}
\usepackage{listings}
\DeclareCaptionStyle{ruled}{labelfont=normalfont,labelsep=colon,strut=off} 
\floatstyle{ruled}
\newfloat{listing}{tb}{lst}{}
\floatname{listing}{Listing}
\title{Low-resource Personal Attribute Prediction from Conversations}
\author {
    Yinan Liu\textsuperscript{\rm 1},
    Hu Chen\textsuperscript{\rm 1},
    Wei Shen\textsuperscript{\rm 1}\thanks{Wei Shen is the corresponding author.},
	Jiaoyan Chen\textsuperscript{\rm 2}
}

\affiliations {
    \textsuperscript{\rm 1} TKLNDST, College of Computer Science, Nankai University, Tianjin 300350, China\\
    \textsuperscript{\rm 2} Department of Computer Science, The University of Manchester\\
    \{liuyn,2120210473\}@mail.nankai.edu.cn, shenwei@nankai.edu.cn, jiaoyan.chen@manchester.ac.uk
}

\begin{document}
	\maketitle

	\begin{abstract}
		Personal knowledge bases (PKBs) are crucial for a broad range of applications such as personalized recommendation and Web-based chatbots. A critical challenge to build PKBs is extracting personal attribute knowledge from users' conversation data. Given some users of a conversational system, a personal attribute and these users' utterances, our goal is to predict the ranking of the given personal attribute values for each user. Previous studies often rely on a relative number of resources such as labeled utterances and external data, yet the attribute knowledge embedded in unlabeled utterances is underutilized and their performance of predicting some difficult personal attributes is still unsatisfactory. In addition, it is found that some text classification methods could be employed to resolve this task directly. However, they also perform not well over those difficult personal attributes. In this paper, we propose a novel framework PEARL to predict personal attributes from conversations by leveraging the abundant personal attribute knowledge from utterances under a low-resource setting in which no labeled utterances or external data are utilized. PEARL combines the biterm semantic information with the word co-occurrence information seamlessly via employing the updated prior attribute knowledge to refine the biterm topic model's Gibbs sampling process in an iterative manner. The extensive experimental results show that PEARL outperforms all the baseline methods not only on the task of personal attribute prediction from conversations over two data sets, but also on the more general weakly supervised text classification task over one data set.
	\end{abstract}
	
	\section{Introduction}
	Personal knowledge bases (PKBs) \cite{balog2019personal,yen2019personal}-structured information about entities personally related to the users, their attributes, and the relations between them-are popular nowadays. They can supply plentiful personal background knowledge for a broad range of downstream applications like Web-based chatbots \cite{ghazvininejad2018knowledge}, personalized recommendation \cite{balog2019transparent,luo2019learning}, and personalized search \cite{lu2020knowledge}. A potential resource for building such PKBs is the personal attribute knowledge (e.g., hobbies and medical conditions) extracted from users' conversation data on a lot of platforms such as social media. To draw the personal attribute knowledge embedded in conversations, personal attribute prediction from conversations becomes an increasingly important task. 
	
	Given multiple users of a conversational system, a personal attribute, and utterances of these users, the task of personal attribute prediction from conversations aims to predict the ranking of the given personal attribute values for each user. It is noted that this task focuses on the case that the personal attribute values are not explicitly mentioned in utterances, and the given attribute values are ranked based on the underlying semantics of utterances, which is different from the common information extraction task. For example, we could rank the attribute values scientist and teacher high with regard to the profession attribute when the user mentions the words ``research", ``lab", ``teaching" and ``educator" in user utterances. However, this task is challenging due to the following aspects: (1) compared with formal documents, user utterances are often short, noisy, colloquial and have diverse topics, and the textual cues in utterances are too implicit to seize; (2) the construction of training data via manually annotating user utterances is time-consuming and labor-intensive; (3) for the personal attribute with too many attribute values (e.g., profession), its several attribute values (e.g., student and teacher) may be related and difficult to distinguish as they often co-occur with the same words in utterances.
	
	Recently, some neural network based models have been explored for this task. These models resort to labeled utterances \cite{tigunova2019listening}, external data (e.g., Wikipedia and Web pages) \cite{liu2022personal} or both \cite{tigunova2020charm} as resources of training data. However, there exist three issues in these previous works: (1) they rely on many resources of training data but these resources are not always available and expensive to fetch, which limits their adaptability to new domains or new data; (2) the attribute knowledge embedded in the unlabeled utterances is underutilized; (3) their performance over some difficult personal attributes (e.g., profession and hobby) is still unsatisfactory. Additionally, it is found that text classification methods \cite{mekala2020contextualized,wang2021x,zhang2021weakly} which are adept in mining the textual cues could be used to address this task directly. Unfortunately, they also fail to achieve good performance when predicting those difficult personal attributes, which has been verified by our experiments.

	Intuitively, the personal attribute knowledge involved in unlabeled utterances is abundant. For example, if the words “law”, “legal”, “court”, and “constitution” frequently co-occur with each other in different utterances, there is a high probability that users who mention these words have the same profession (i.e., lawyer). This kind of word co-occurrence information is beneficial for predicting personal attributes. Moreover, it is found that different word pairs (biterms) constructed from utterances as well as two words belonging to the same biterm may be related to an attribute value in different degrees at the semantic level. This kind of biterm semantic information is attribute-oriented and is also vital to our task. Consequently, how to integrate these two categories of information becomes very crucial.
	
	To tackle the above issues, we propose a novel framework PEARL to \textbf{\underline{P}}redict p\textbf{\underline{E}}rsonal \textbf{\underline{A}}ttributes from conve\textbf{\underline{R}}sations by leveraging the abundant personal attribute knowledge from utterances in a \textbf{\underline{L}}ow-resource setting (without requiring any labeled utterances or external data). Our proposed framework PEARL is composed of a biterm semantic acquisition (BSA) module and an attribute knowledge integration (AKI) module. To capture the biterm semantic information, the BSA module derives the biterm set by searching words with high semantic relevance to the attribute value from utterances and yields the attribute-oriented biterm representation for each biterm based on the pre-trained language model (PLM). To integrate the biterm semantic information with the word co-occurrence information, the AKI module leverages the biterm-attribute value similarity score derived from the BSA module as the prior attribute knowledge to guide the biterm topic model (BTM)'s \cite{yan2013biterm} Gibbs sampling process. To further promote the prediction performance, the AKI module can update the prior attribute knowledge based on the new sampling result and refine the Gibbs sampling process guided by the updated prior attribute knowledge in an iterative manner. After multiple iterations in such a way, PEARL can predict probable attribute values for each user by utilizing the final sampling result. Additionally, it is worth mentioning that our proposed framework can not only solve the personal attribute prediction task but also adapt to the more general weakly supervised text classification task.
	
	Our contributions can be summarized as follows:
	\begin{itemize}
		\item We are the first to address personal attribute prediction from conversations under a low-resource setting which does not resort to any labeled utterances or external data.
		\item We propose a novel framework PEARL which fuses the biterm semantic information and the word co-occurrence information together via leveraging the updated prior attribute knowledge to refine the BTM's Gibbs sampling process in an iterative manner. 
		\item Extensive experimental studies have been conducted for the task of personal attribute prediction from conversations over two data sets, and the task of weakly supervised text classification over one data set. The experimental results show that our framework surpasses all the baseline methods on both tasks. 
	\end{itemize}

	\section{Preliminary}
	\subsection{Task Definition}
	
	A user ID of a conversational system is denoted by $i$ ($1\le i\le n$), and the utterances posted by this user can be concatenated as one utterance denoted by $u_i$. A personal attribute is denoted by $c$, and $\{c_1, c_2, ..., c_g\}$ is the set of personal attribute values of $c$. Formally, given user IDs $1, 2, ..., n$, a personal attribute $c$, and the corresponding utterances of these users, the task of low-resource personal attribute prediction from conversations is to predict the ranking of the personal attribute values (i.e., $c_{1}, c_{2},..., c_{g}$) for each user according to these unlabeled utterances without the need of any other resources (e.g., labeled utterances or external data).

	\subsection{Biterm Topic Model} Conventional topic models (e.g., PLSA \cite{hofmann1999probabilistic} and LDA \cite{blei2003latent}) can reveal the latent topics by implicitly capturing the document-level word co-occurrence patterns, while biterm topic model (BTM) can learn the topics better by modeling the generation of word co-occurrence patterns directly. We introduce BTM here in brief. Given a biterm set $B$, which is constructed based on a document collection by extracting any two distinct words in a document as a biterm, the generation process of $B$ in BTM can be described as follows:
	\begin{itemize}
		\item Draw a topic distribution $\theta\sim \rm{Dir(\alpha)}$
		\item For each topic $z$
		
		(a) Draw a topic-word distribution $\psi_z\sim \rm{Dir(\beta)}$

		\item For each biterm $b \in B$
		
		(a) Draw a topic $z \sim \rm{Multi(\theta)}$
		
		(b) Draw two words to generate $b$: $w_j$,$w_k$$\sim \rm{Multi(\psi_z)}$ 
	\end{itemize}
	where $\alpha$ and $\beta$ are the Dirichlet priors. The parameters of BTM (i.e., $\psi$ and $\theta$) can be approximately inferred by the Gibbs sampling process. To perform Gibbs sampling, the key step is to calculate the conditional distribution for each biterm $b$ as follows:
	\begin{equation}
	\begin{split}
	P(z|\textbf{z}_{\neg {b}},B,\alpha,\beta)\propto (n_z+\alpha)
	\cdot\frac{(n_{w_j|z}+\beta)(n_{w_k|z}+\beta)}{(\sum_{w}n_{w|z}+M\beta)^2}
	\label{pzbi}
	\end{split}
	\end{equation}
	where $\textbf{z}_{\neg {b}}$ denotes the topic assignments for all biterms except $b$; $n_{w|z}$ denotes how many times the word $w$ is assigned to the topic $z$; $n_z$ denotes how many biterms are assigned to the topic $z$; $M$ denotes the number of words in the document collection. Subsequently, the parameters $\psi$ and $\theta$ can be calculated as follows:
	\begin{equation}
	\psi_{w|z}=\frac{n_{w|z}+\beta}{\sum_{w}n_{w|z}+M\beta}
	\label{psig}
	\end{equation}
	\begin{equation}
	\theta_{z}=\frac{n_{z}+\alpha}{|B|+J\alpha}
	\label{thetag}
	\end{equation}
	where $J$ is the number of topics. Finally, the topic proportions of a document $d$ can be inferred based on $\psi$ and $\theta$ as follows:
	\begin{equation}
	P(z|d)=\sum_{b}P(z|b)P(b|d)
	\label{pzd}
	\end{equation}
	\begin{equation}
	\hspace{-1mm}P(z|b)\!=\!\frac{P(z)P(w_j|z)P(w_k|z)}{\sum_{z}P(z)P(w_j|z)P(w_k|z)}\!=\!\frac{\theta_{z}\psi_{w_j|z}\psi_{w_k|z}}{\sum_{z}\theta_{z}\psi_{w_j|z}\psi_{w_k|z}}
\label{pzb}
	\end{equation}
	where $P(b|d)$ denotes the relative frequency of $b$ in $d$. Interested readers please refer to BTM \cite{yan2013biterm} for more details. 
	
	\section{The Framework PEARL}
\begin{figure}[!t]
		\centering
		\includegraphics[width=0.46\textwidth]{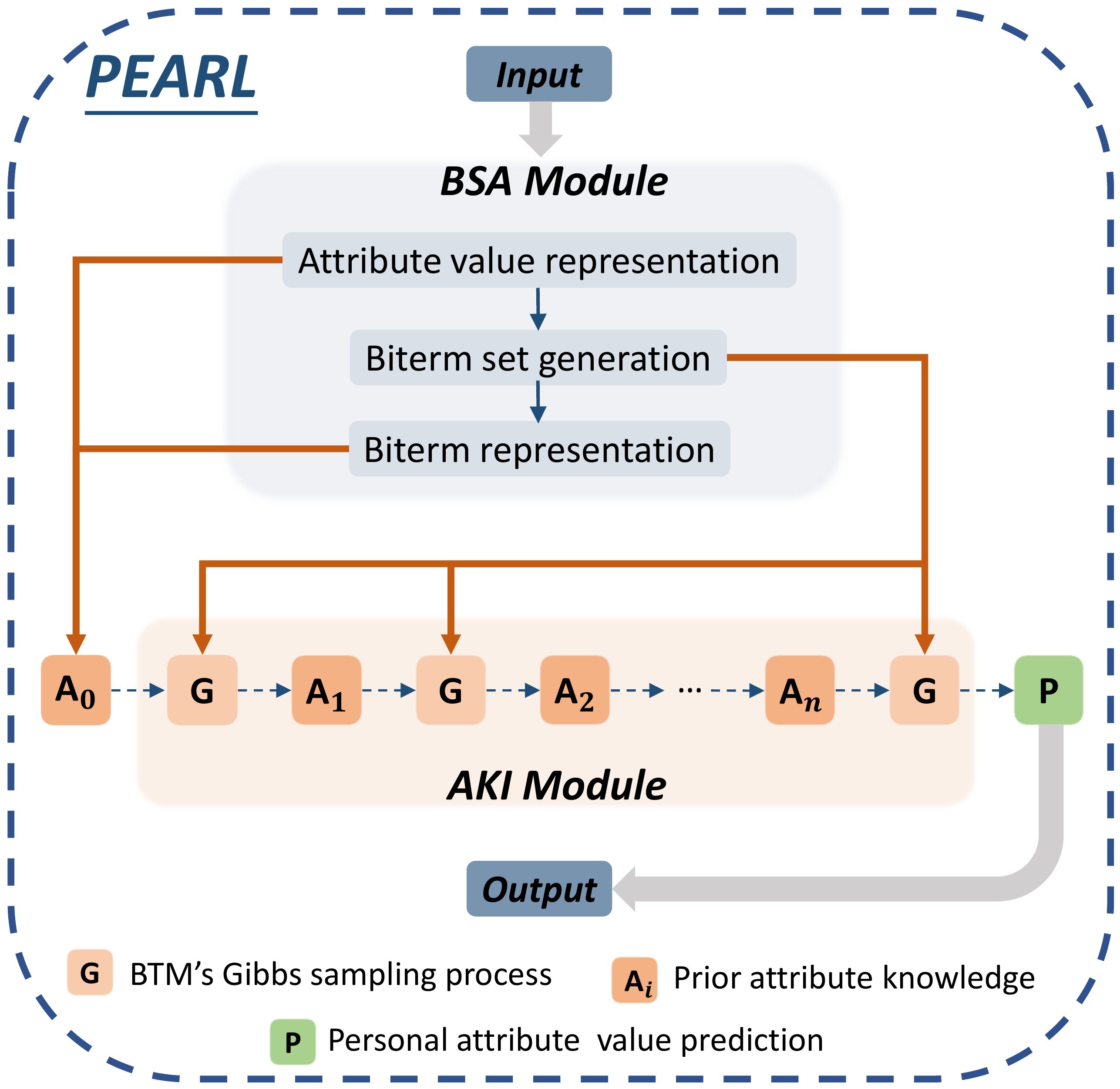}
		\caption{Overview of our framework PEARL.}
		\label{framework}
	\end{figure}

	The overall framework of our proposed PEARL is shown in Figure \ref{framework}. We begin with the introduction of the BSA module and thereafter describe the AKI module.
	\subsection{Biterm Semantic Acquisition (BSA) Module} 
	The BSA module consists of three parts: (1) attribute value representation (i.e., construct an attribute value representation for each attribute value); (2) biterm set generation (i.e., generate a biterm set for each utterance based on the attribute value representation); (3) biterm representation (i.e., yield an attribute-oriented biterm representation for each biterm based on the biterm set). We elaborate them as follows.
	\subsubsection{Attribute Value Representation.} To understand the semantics of personal attribute values, we first define the static representation $s_w$ for a word $w$ by averaging contextualized representations of its all occurrences in utterances as follows:
	\begin{equation}
	s_w=\frac{\sum_{u_{i,j}=w}r_{i,j}}{N_w}
	\label{sw}
	\end{equation}
	where $u_{i,j}$ denotes the $j$-th token of utterance $u_i$; $r_{i,j}$ denotes the contextualized token representation of $u_{i,j}$ based on a PLM; $N_w$ denotes how many times $w$ appears in all utterances. Thus, for an attribute value $c_q$ which contains only one word, its static representation can be denoted by $s_{c_q}$. Specially, if $c_q$ contains several words, we average the static representations of all the single words in $c_q$ as $s_{c_q}$ for simplicity.
	
	To enhance the semantic understanding, inspired by recent weakly supervised text classification methods \cite{wang2021x,mekala2020contextualized}, we propose to construct a word list for each attribute value to store some attribute-related words. For the attribute value $c_q$, we utilize its surface form to initialize its word list $L_{c_q}$, and subsequently discover its next attribute-related word owning the highest similarity score in utterances to expand its current word list iteratively. This similarity score can be calculated by the cosine similarity between the static representation of a word and the static representation of an attribute value. If the selected word is in the current word list of an attribute value, we will ignore it and continue to select another word for this iteration until the selected word does not occur in any attribute value's current word list. We will stop the iteration for the attribute value $c_q$ if the overlap between the current word list $L_{c_q}$ and the top $|L_{c_q}|$ similar words generated by the current attribute value representation is below a threshold $\eta$, which is a dynamic mechanism. We define the current attribute value representation $\nu_{c_q}$ for the attribute value $c_q$ as a weighted average representation based on the current word list $L_{c_q}$ of $c_q$ as follows:	
	\begin{equation}
	\nu_{c_q}=\sum_{i=1}^{|L_{c_q}|}f(s_{L_{{c_q},i}}, s_{c_q})\cdot s_{L_{{c_q},i}}
	\end{equation}
	where ${L_{{c_q},i}}$ denotes the $i$-th word in ${L_{{c_q}}}$ ($1\le i\le |L_{c_q}|$) and $s_{L_{{c_q},i}}$ denotes the static representation of $L_{{c_q},i}$ calculated by Formula \ref{sw}; $f(s_{L_{{c_q},i}}, s_{c_q})$ is the normalized cosine similarity score between $s_{L_{{c_q},i}}$ and $s_{c_q}$. It is also worth mentioning that the final lengths of the word lists for different attribute values could be different as they are determined by the dynamic mechanism.

	\subsubsection{Biterm Set Generation.} Intuitively, different words in an utterance have various levels of importance for predicting personal attributes. Specially, those words that are highly relevant to the attribute value are valuable for our task, so we need to find them out from each utterance to compose the biterm set. Inspired by the previous work \cite{xie2016unsupervised}, to estimate the word importance, firstly we calculate the weight $\pi_{i,j}$ for each word $u_{i,j}$ based on the Student t-distribution \cite{van2008visualizing} via using the obtained attribute value representation as follows:
	\begin{equation}
	\pi_{i,j}=\mathop{\rm{max}}\limits_{c_q}Sim(u_{i,j},c_q)
	\end{equation}
	\begin{equation}
	\label{simuc}
	Sim(u_{i,j},c_q)=\frac{(1\ +\ \rVert r_{i,j}-\nu_{c_q}\rVert^{2}/\lambda)^{-\frac{\lambda+1}{2}}}{\sum_{c'_q}(1\ + \ \rVert r_{i,j}-\nu_{c'_q}\rVert^{2}/\lambda)^{-\frac{\lambda+1}{2}}}
	\end{equation}
	where $Sim(u_{i,j},c_q)$ is the similarity between the word $u_{i,j}$ and the attribute value $c_q$; $\lambda$ is the degree of freedom for the Student t-distribution. Subsequently, we aims to select top $K$ words with high weights from each utterance $u_i$ as the keywords, i.e., $u_{i,top_1}$, $u_{i,top_2}$,... , $u_{i,top_h}$, ..., and $u_{i,top_K}$, where $1\le h\le K$ and $top_h$ is the ID of the token with the $h$-th highest weight in $u_i$. Thus we can construct a biterm set $\mathfrak{B}_i$ for $u_i$ via the selected $K$ keywords of $u_i$ as follows:
	\begin{equation}
	\mathfrak{B}_i=\{b_{i,h,l}=(u_{i,top_h},u_{i,top_l})|1\le h,l\le K\}\ \ 		
	\end{equation} where $b_{i,h,l}$ denotes a biterm composed of the words $u_{i,top_h}$ and $u_{i,top_l}$. By concatenating the biterm sets $\mathfrak{B}_1, \mathfrak{B}_2,...,\mathfrak{B}_n$ we can obtain the final biterm set $\mathfrak{B}$ over all utterances for our task. 

	\subsubsection{Biterm Representation.} Considering that the two words belonging to the same biterm have different relevance degrees to the attribute value at the semantic level, we propose to construct the attribute-oriented biterm representation for each biterm to capture the biterm semantic information. After obtaining the biterm set $\mathfrak{B}$, for each biterm $b_{i,h,l}$ in $\mathfrak{B}$, we define its representation $\nu_{b_{i,h,l}}$ as follows:
	\begin{equation}
	\nu_{b_{i,h,l}}=\pi_{i,top_h}r_{i,top_h}+\pi_{i,top_l}r_{i,top_l}
	\end{equation}
	It is noted that for the utterance $u_i$, the weights of its selected $K$ keywords should be normalized in advance. 

	\subsection{Attribute Knowledge Integration (AKI) Module} 
	\label{aki}
	Although biterm topic model (BTM) exploits the word co-occurrence information successfully, the biterm semantic information is ignored in BTM. To fuse these two categories of information, we try to inject the biterm semantic information into BTM via this AKI module. 
	If we run BTM directly on the biterm set $\mathfrak{B}$ generated by the BSA module, i.e., iteratively calculate the conditional distribution $P(z|\textbf{z}_{\neg {b}},B,\alpha,\beta)$ for each biterm (Formula \ref{pzbi}) and update the counting variables (i.e., $n_z$ and $n_{w|z}$), each utterance can be assigned a topic by calculating the highest value of $P(z|u_i)$ (Formula \ref{pzd}) over all the topics. However, even if the number of topics is set to be the same as the number of attribute values,  the correspondence between the topic and the attribute value is still lacking, which makes it impossible to be applied to our task of personal attribute prediction. 
	
	To remedy this issue, we propose an iterative biterm semantics (BS) based Gibbs sampling process in the AKI module. It is able to build the corresponding relationship between the attribute value and the topic via leveraging the prior attribute knowledge derived from the BSA module to guide the BTM's Gibbs sampling process. First, we associate each topic $z$ with an individual attribute value $c_q$, and initialize states for the Markov chain randomly like BTM. Next, inspired by \cite{yang2021dataless}, we define the conditional distribution $P'(c_{q}|\textbf{c}_{\neg {b_{i,h,l}}},\mathfrak{B},\alpha,\beta)$ for each biterm $b_{i,h,l} $ in the biterm set $\mathfrak{B}$ via combining the biterm-attribute value similarity score $\Omega(b_{i,h,l},c_q)$ with the conditional distribution $P(c_{q}|\textbf{c}_{\neg {b_{i,h,l}}},\mathfrak{B},\alpha,\beta)$ (Formula \ref{pzbi}) as follows:
	\begin{equation}
	\begin{split}
	P'(c_{q}|\textbf{c}_{\neg {b_{i,h,l}}},\mathfrak{B},\alpha,\beta)=  \Omega(b_{i,h,l},c_q)\\
	\cdot  P(c_{q}|\textbf{c}_{\neg {b_{i,h,l}}},\mathfrak{B},\alpha,\beta)
	\end{split}
	\label{pos}
	\end{equation}
	where $\textbf{c}_{\neg {b_{i,h,l}}}$ denotes the the attribute value assignments for all
	biterms of $\mathfrak{B}$ except $b_{i,h,l}$.
	The biterm-attribute value similarity score could encode the prior attribute knowledge involved in utterances well, and we initialize it using the value of $cosine(\nu_{b_{i,h,l}},\nu_{c_q})$, where $\nu_{b_{i,h,l}}$ and $\nu_{c_q}$ denote the representations of the biterm $b_{i,h,l}$ and the attribute value $c_q$ respectively, which are yielded by the BSA module. Following the original Gibbs sampling process, we iteratively calculate the conditional distribution for each biterm by Formula \ref{pos} and update the counting variables $n_{c_q}$, $n_{u_{i,top_h}|c_{q}}$, and $n_{u_{i,top_l}|c_{q}}$. Finally, the parameters $\psi$ and $\theta$ of BTM can be calculated by Formulas \ref{psig} and \ref{thetag} respectively. In this way, the AKI module enables BTM to induce attribute value-aware topics in the inference stage to integrate the biterm semantic information and the word co-occurrence information preliminarily. To further fuse these two categories of information, we propose a simple iteration operation by exploiting superior prior attribute knowledge to refine the Gibbs sampling process. Specifically, we update $\Omega(b_{i,h,l},c_q)$ according to the parameters $\psi$ in an iterative manner, which are the output of the Gibbs sampling process for each iteration as follows:
	\begin{equation}
	\Omega(b_{i,h,l},c_q)= \psi_{u_{i,top_h}|c_q}\cdot \psi_{u_{i,top_l}|c_q}
	\label{priu}	
	\end{equation} 
It is noted that this iteration operation can promote the performance of our framework PEARL successfully, which has been verified in our experiments. In practice, it is found that after dozens of iterations, the performance is stable. This iterative procedure is summarized in Algorithm \ref{a1}, which can automatically learn the parameters of our framework PEARL without requiring any training data. Note that in Algorithm 1, $E$ is the number of iterations for the proposed iteration operation, while $T$ is the number of iterations for the Gibbs sampling process.

	\begin{algorithm}[t!]
		\caption{$\rm {Iterative \ BS \ based \ Gibbs \ sampling\ process}$}
		\label{a1}
		\textbf{Input:} The biterm set $\mathfrak{B}$, the biterm representation $\nu_{b_{i,h,l}}$ of the biterm $b_{i,h,l}\in\mathfrak{B}$, the attribute value representations $\nu_{c_{1}}$, $\nu_{c_{2}}$, ..., $\nu_{c_{g}}$,  hyperparameters $\alpha$ and $\beta$, and the number of iterations $E$ and $T$.\\
		\textbf{Output:} The parameters $\psi$ and $\theta$.
		\begin{algorithmic}[1]
			
			\STATE Initialize attribute value assignments randomly for all the
			biterms in $\mathfrak{B}$
			\STATE Initialize the biterm-attribute value similarity score $\Omega(b_{i,h,l},c_{q})$ by the value of $cosine(\nu_{b_{i,h,l}},\nu_{c_{q}})$
			\FOR{$e=1$ to $E$ }
			\FOR{$t=1$ to $T$ }  
			
			\FOR{$b_{i,h,l} \in \mathfrak{B}$} 
			\STATE Draw an attribute value for ${b_{i,h,l}}$ from the conditional distribution $P'(c_q|\textbf{c}_{\neg {b_{i,h,l}}},\mathfrak{B},\alpha,\beta)$ by Formula \ref{pos}
			\STATE Update $n_{c_q}$, $n_{u_{i,top_h}|c_{q}}$, and $n_{u_{i,top_l}|c_{q}}$
			\ENDFOR
			\ENDFOR
			\STATE Calculate $\psi$ and $\theta$ by Formulas \ref{psig} and \ref{thetag} respectively
			\STATE Update $\Omega(b_{i,h,l},c_q)$ by Formula \ref{priu}
			\ENDFOR 
		\end{algorithmic}
	\end{algorithm}
	After performing the iterative BS based Gibbs sampling process, the ranking of the attribute values can be obtained based on the learned parameters $\psi$ and $\theta$ via calculating the probability score $P(c_q|u_i)$ (Formula \ref{pzd}) for each attribute value $c_q$. 
	

	\begin{table*}[!t]
		\centering
		
		\resizebox{0.85\textwidth}{!}{
			\begin{tabular}{|c|c|cc|c|r|r|r|r|}
				\hline
				\multirow{2}{*}{\textbf{\textit{Method type}}} & {\textbf{\textit{Labeled}}} & \multicolumn{2}{c|}{\textbf{\textit{External data}}} & \multirow{2}{*}{\textbf{\textit{Method}}} & \multicolumn{2}{c|}{\textbf{\textit{Profession}}} & \multicolumn{2}{c|}{\textbf{\textit{Hobby}}} \\
				\cline{3-4}\cline{6-9}          &   \textbf{ \textit{utterances}}    & \multicolumn{1}{l}{\textbf{\textit{Wiki-page}}} & \multicolumn{1}{l|}{\textbf{\textit{Wiki-category}}} &       & \multicolumn{1}{c}{\textbf{\textit{MRR}}} & \multicolumn{1}{c|}{\textbf{\textit{nDCG}}} & \multicolumn{1}{c}{\textbf{\textit{MRR}}} & \multicolumn{1}{c|}{\textbf{\textit{nDCG}}} \\
				\hline
				\hline
				\multirow{26}{*}{\makecell[c]{Personal attribute\\ prediction\\ methods 
				}}
				& \multirow{6}{*}{yes} & \multicolumn{1}{c}{\checked} &x       & \multirow{2}{*}{{BERT IR (2019)}} & \multicolumn{1}{r}{0.30} & \multicolumn{1}{r|}{0.45} & \multicolumn{1}{r}{0.22} & 0.43 \\
			         &       & x     & \checked     &       & \multicolumn{1}{r}{0.28} & \multicolumn{1}{r|}{0.44} & \multicolumn{1}{r}{0.18} & 0.42 \\
				\cline{3-9}          &       & \multicolumn{1}{c}{\checked} & x     & \multirow{2}{*}{$\rm{CHARM_{KNRM}}$ (2020)} & \multicolumn{1}{r}{0.27} & \multicolumn{1}{r|}{0.44} & \multicolumn{1}{r}{0.22} & 0.44 \\
				         &       & x     & \checked     &       & \multicolumn{1}{r}{0.35} & \multicolumn{1}{r|}{0.55} & \multicolumn{1}{r}{0.27} & 0.49 \\
				\cline{3-9}          &       & \multicolumn{1}{c}{\checked} & x     & \multirow{2}{*}{${\rm{CHARM_{BM25}}}$ (2020)} & \multicolumn{1}{r}{0.29} & \multicolumn{1}{r|}{0.46} & \multicolumn{1}{r}{0.24} & 0.47 \\
				         &       & x     & \checked     &       & \multicolumn{1}{r}{0.28} & \multicolumn{1}{r|}{0.47} & \multicolumn{1}{r}{0.21} & 0.43 \\
				\Xcline{2-9}{1pt}
				& \multirow{26}{*}{no} & \multicolumn{1}{c}{\checked} & x    & \multirow{2}{*}{No-keyword + BM25 (1995)} & \multicolumn{1}{r}{0.15} & \multicolumn{1}{r|}{0.32} & \multicolumn{1}{r}{0.16} & 0.42 \\
			       &       &x     &\checked     &       & \multicolumn{1}{r}{0.17} & \multicolumn{1}{r|}{0.37} & \multicolumn{1}{r}{0.13} & 0.35 \\
				\cline{3-9}          &       & \multicolumn{1}{c}{\checked} & x    & \multirow{2}{*}{RAKE + BM25 (2010, 1995)} & \multicolumn{1}{r}{0.16} & \multicolumn{1}{r|}{0.33} & \multicolumn{1}{r}{0.17} & 0.42 \\
			         &       & x     & \checked     &       & \multicolumn{1}{r}{0.19} & \multicolumn{1}{r|}{0.39} & \multicolumn{1}{r}{0.14} & 0.37 \\
				\cline{3-9}          &       & \multicolumn{1}{c}{\checked} & x    & \multirow{2}{*}{RAKE + KNRM (2010, 2017)} & \multicolumn{1}{r}{0.16} & \multicolumn{1}{r|}{0.33} & \multicolumn{1}{r}{0.12} & 0.32 \\
			         &       & x     & \checked     &       & \multicolumn{1}{r}{0.13} & \multicolumn{1}{r|}{0.34} & \multicolumn{1}{r}{0.12} & 0.31 \\
				\cline{3-9}          &       & \multicolumn{1}{c}{\checked} &  x   & \multirow{2}{*}{TextRank + BM25 (2004, 1995)} & \multicolumn{1}{r}{0.21} & \multicolumn{1}{r|}{0.39} & \multicolumn{1}{r}{0.21} & 0.46 \\
			         &       &x    & \checked    &       & \multicolumn{1}{r}{0.26} & \multicolumn{1}{r|}{0.45} & \multicolumn{1}{r}{0.20} & 0.42 \\
				\cline{3-9}          &       & \checked     & x     & \multirow{2}{*}{TextRank + KNRM (2004, 2017)} & \multicolumn{1}{r}{0.21} & \multicolumn{1}{r|}{0.38} & \multicolumn{1}{r}{0.15} & 0.36 \\
			        &       & x   & \checked     &       & \multicolumn{1}{r}{0.18} & \multicolumn{1}{r|}{0.36} & \multicolumn{1}{r}{0.16} & 0.36 \\
				\cline{3-9}          &       & \multicolumn{1}{c}{\checked} & x  & \multirow{2}{*}{$\rm{{HAM}_{avg}}$ (2019)} & \multicolumn{1}{r}{0.06} & \multicolumn{1}{r|}{0.07} & \multicolumn{1}{r}{0.06} & 0.05 \\
			        &       & x   & \checked     &       & \multicolumn{1}{r}{0.06} & \multicolumn{1}{r|}{0.07} & \multicolumn{1}{r}{0.03} & 0.02 \\
				\cline{3-9}          &       & \multicolumn{1}{c}{\checked} &  x  & \multirow{2}{*}{${\rm{HAM}_{2attn}}$ (2019)} & \multicolumn{1}{r}{0.06} & \multicolumn{1}{r|}{0.07} & \multicolumn{1}{r}{0.04} & 0.05 \\
			        &       & x     & \checked     &       & \multicolumn{1}{r}{0.06} & \multicolumn{1}{r|}{0.07} & \multicolumn{1}{r}{0.06} & 0.07 \\
				\cline{3-9}          &       & \multicolumn{1}{c}{\checked} & x    & \multirow{2}{*}{${\rm{HAM}_{CNN}}$ (2019)} & \multicolumn{1}{r}{0.20} & \multicolumn{1}{r|}{0.18} & \multicolumn{1}{r}{0.22} & 0.14 \\
			         &       & x     & \checked     &       & \multicolumn{1}{r}{0.27} & \multicolumn{1}{r|}{0.34} & \multicolumn{1}{r}{0.17} & 0.27 \\
				\cline{3-9}          &       & \multicolumn{1}{c}{\checked} & x     & \multirow{2}{*}{${\rm{HAM}_{CNN-attn}}$ (2019)} & \multicolumn{1}{r}{0.21} & \multicolumn{1}{r|}{0.28} & \multicolumn{1}{r}{0.13} & 0.10 \\
		         &       & x     & \checked     &       & \multicolumn{1}{r}{0.25} & \multicolumn{1}{r|}{0.31} & \multicolumn{1}{r}{0.16} & 0.25 \\
				\cline{3-9}          &       & \multicolumn{1}{c}{\checked} & x    & \multirow{2}{*}{ DSCGN (2022)} & \multicolumn{1}{r}{0.43} & \multicolumn{1}{r|}{0.57} & \multicolumn{1}{r}{0.29} & 0.50 \\
			         &       & x     & \checked     &       & \multicolumn{1}{r}{0.44} & \multicolumn{1}{r|}{0.60} & \multicolumn{1}{r}{0.29} & 0.49 \\
				
				\Xcline{1-9}{1pt}
				
				\multirow{5}{*}{\makecell[c]{Weakly supervised\\ text classification\\ methods}} & \multicolumn{1}{c|}{\multirow{5}{*}{no}} & x     & x     & \multirow{1}{*}{\makecell[c]{SeedBTM (2020)}} & \multicolumn{1}{r}{0.33} & \multicolumn{1}{r|}{0.55} & \multicolumn{1}{r}{0.17} & 0.42 \\
				          &       & x     & x    & \multirow{1}{*}{ConWea (2020)} & \multicolumn{1}{r}{0.07} & \multicolumn{1}{r|}{0.26} & \multicolumn{1}{r}{0.04} & 0.21 \\
			         &       & x    & x     & \multirow{1}{*}{LOTClass (2020)} & \multicolumn{1}{r}{0.07} & \multicolumn{1}{r|}{0.26} & \multicolumn{1}{r}{0.04} & 0.2 \\
			       &       & x     & x    & \multirow{1}{*}{X-Class (2021)} & \multicolumn{1}{r}{0.34} & \multicolumn{1}{r|}{0.57} & \multicolumn{1}{r}{0.23} & 0.47 \\
			        &       & x     & x     & \multirow{1}{*}{ClassKG (2021)} & \multicolumn{1}{r}{0.07} & \multicolumn{1}{r|}{0.24} & \multicolumn{1}{r}{0.04} & 0.19 \\
				\rowcolor{mygray}
				\Xhline{1pt}
				\multirow{1}{*}{Our method} & \multicolumn{1}{c|}{no}& x     & x     & \multirow{1}{*}{PEARL} & \multicolumn{1}{r}{\textbf{0.49}} & \multicolumn{1}{r|}{\textbf{0.64}} & \multicolumn{1}{r}{\textbf{0.31}} & \textbf{0.54} \\
				\hline
		\end{tabular}}%
\caption{Performance on the task of personal attribute prediction from conversations. All the results of the personal attribute prediction baselines are taken from DSCGN \cite{liu2022personal}. The performance of all the weakly supervised text classification methods is reproduced via their open-source solutions.}
		\label{effectiveness}%
	\end{table*}%

	\section{Experiments}

	\subsection{Experimental Setting}
	\noindent\textbf{Data Sets.}
	For the task of personal attribute prediction from conversations, we perform experiments over two public data sets: (1) profession data set; (2) hobby data set. These two data sets are extracted from publicly-available Reddit submissions and comments ($2006$ - $2018$), and are annotated and provided by the authors of \cite{tigunova2020charm}. All utterances containing explicit personal attribute assertions used for annotation have been removed. The given attribute values for each personal attribute are defined based on Wikipedia lists. The number of attribute values for profession and hobby are $71$ and $149$ respectively. Both data sets consist of about $6000$ users and have a maximum of $500$ and an average of $23$ users for each personal attribute value.
	
	\vspace{2 pt}
	
	\noindent\textbf{Evaluation Measures.} Following the previous personal attribute prediction studies \cite{tigunova2019listening,tigunova2020charm,liu2022personal}, we adopt the same ranking metrics MRR (Mean Reciprocal Rank) and nDCG (normalized Discounted Cumulative Gain) to evaluate all the methods. 
	
	\vspace{2 pt}
	
	\noindent\textbf{Setting Details.} The threshold $\eta$, the Dirichlet prior $\beta$, the number of keywords $K$ for each utterance, the degree of freedom $\lambda$, the numbers of iterations $E$ and $T$ are set to $75\%$, $0.01$, $60$, $1$, $20$ and $50$ respectively. The Dirichlet prior $\alpha$ is set to $50/g$, where $g$ is the number of attribute values. BERT base-uncased model is adopted as the PLM. The number of attribute-related words for each attribute value is set to a minimum of $10$ and a maximum of $40$. The experiments are implemented by MindSpore Framework\footnote{https://www.mindspore.cn/en}. The source code and data sets used in this paper are publicly available\footnote{https://github.com/CodingPerson/PEARL}.
	
	\begin{figure*}[htbp]
		\centering
		\begin{subfigure}[t]{0.246\textwidth}
			\centering
			\includegraphics[width=\textwidth]{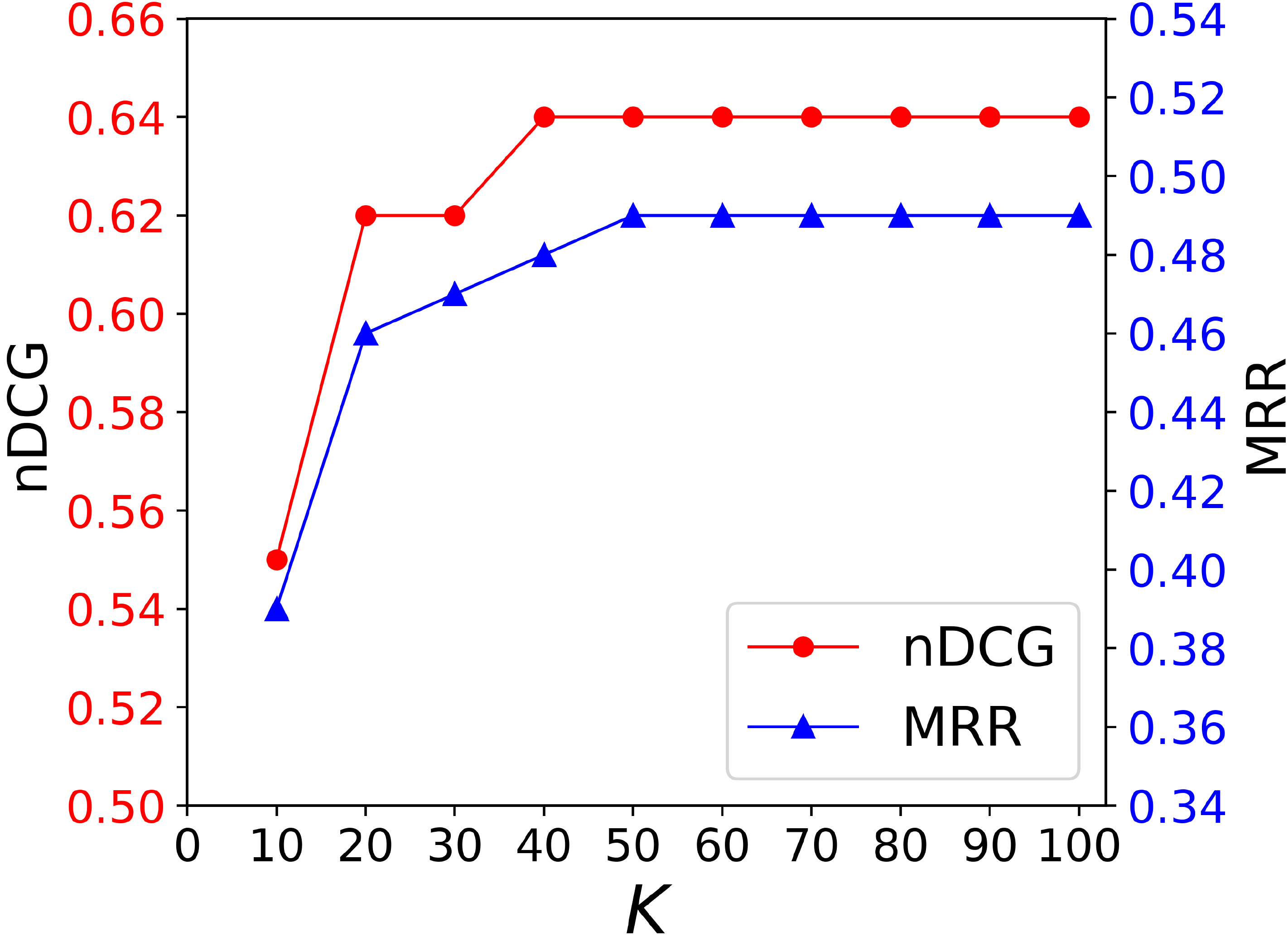}
			\caption{profession, varying $K$.}
			\label{fig:a}
		\end{subfigure}
		\begin{subfigure}[t]{0.246\textwidth}
			\centering
			\includegraphics[width=\textwidth]{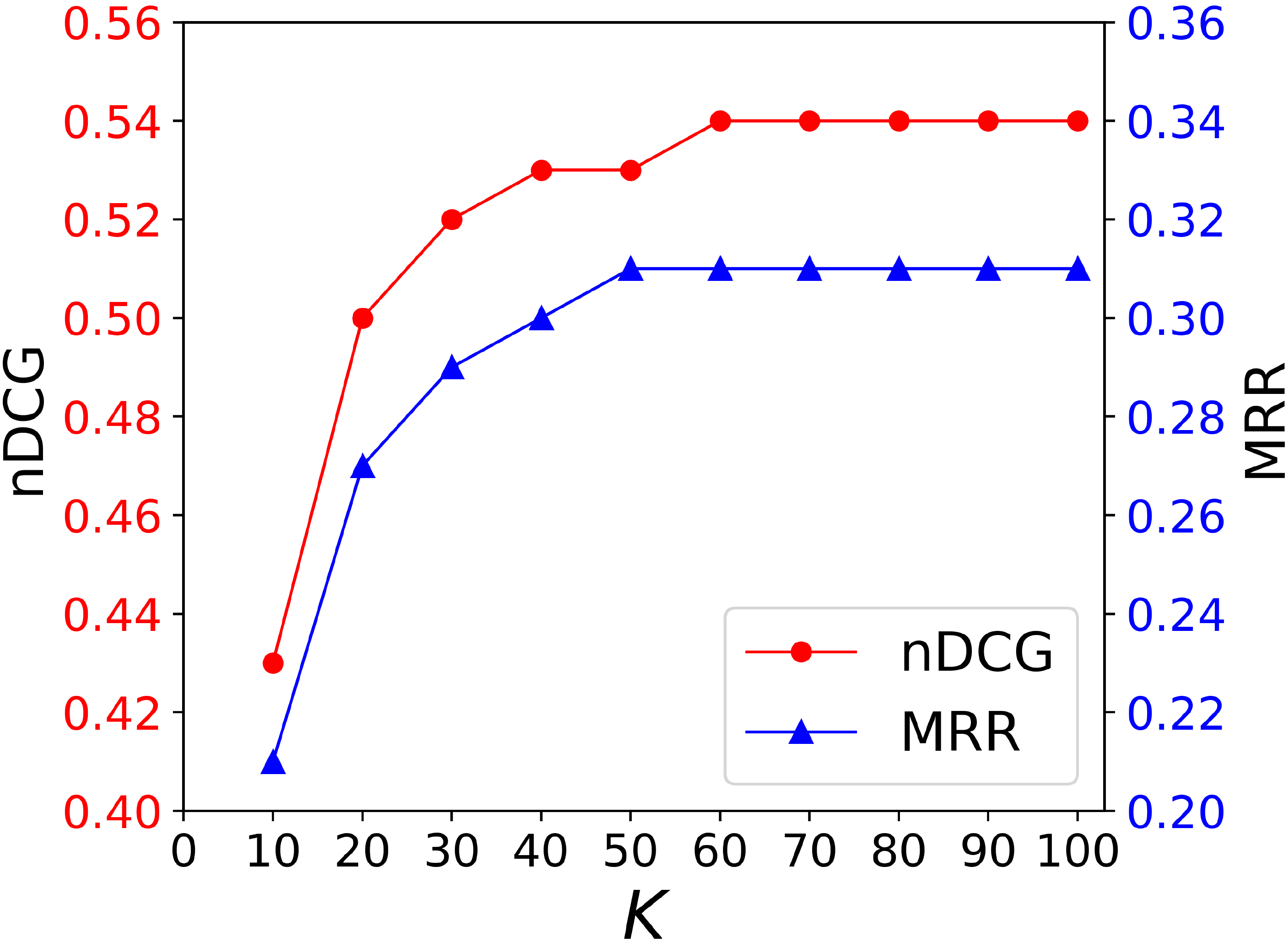}
			\caption{hobby, varying $K$.}
			\label{fig:b}
		\end{subfigure}
		\begin{subfigure}[t]{0.246\textwidth}
			\centering
			\includegraphics[width=\textwidth]{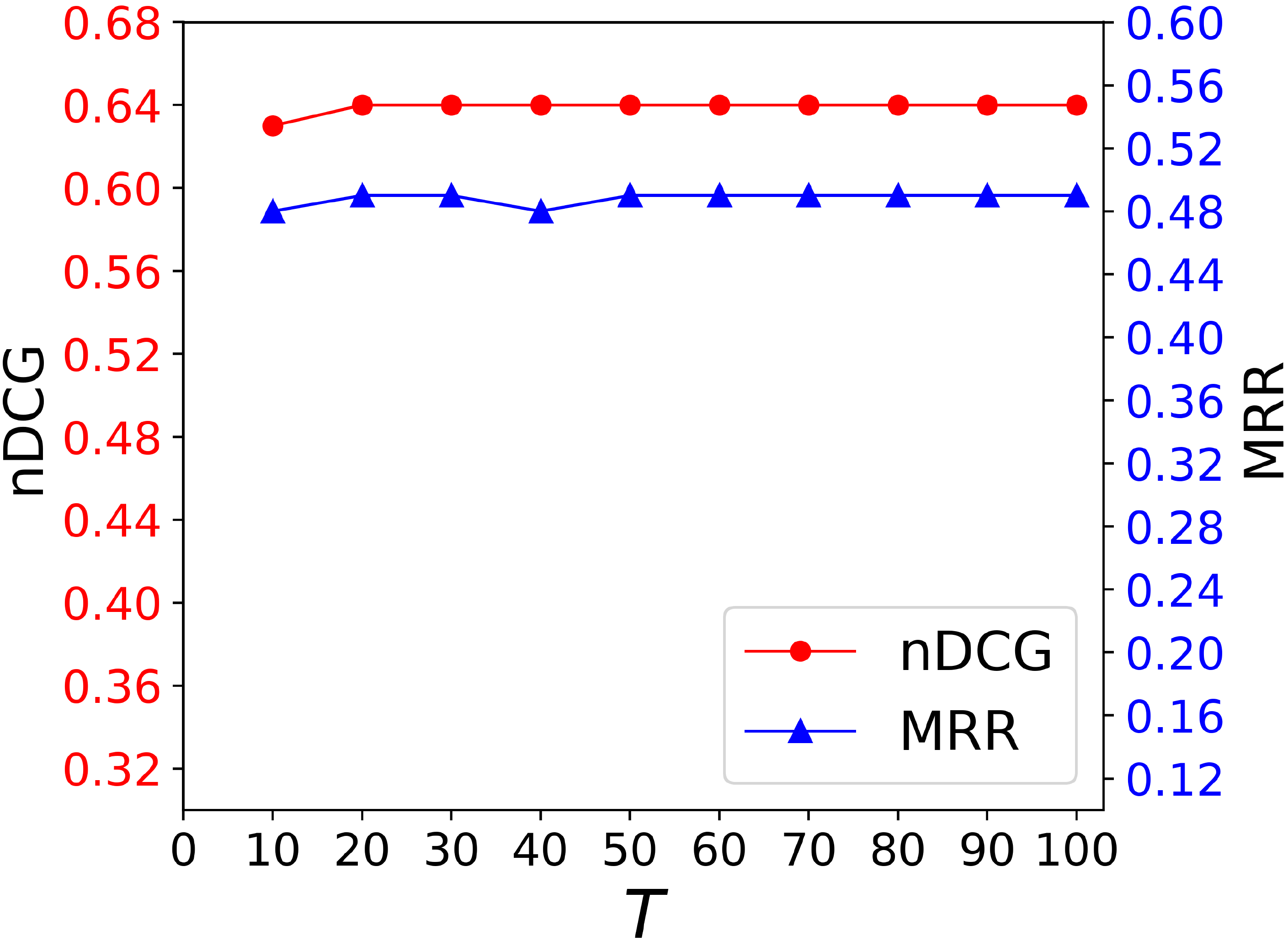}
			\caption{profession, varying $T$.}
			\label{fig:c}
		\end{subfigure}
		\begin{subfigure}[t]{0.246\textwidth}
			\centering
			\includegraphics[width=\textwidth]{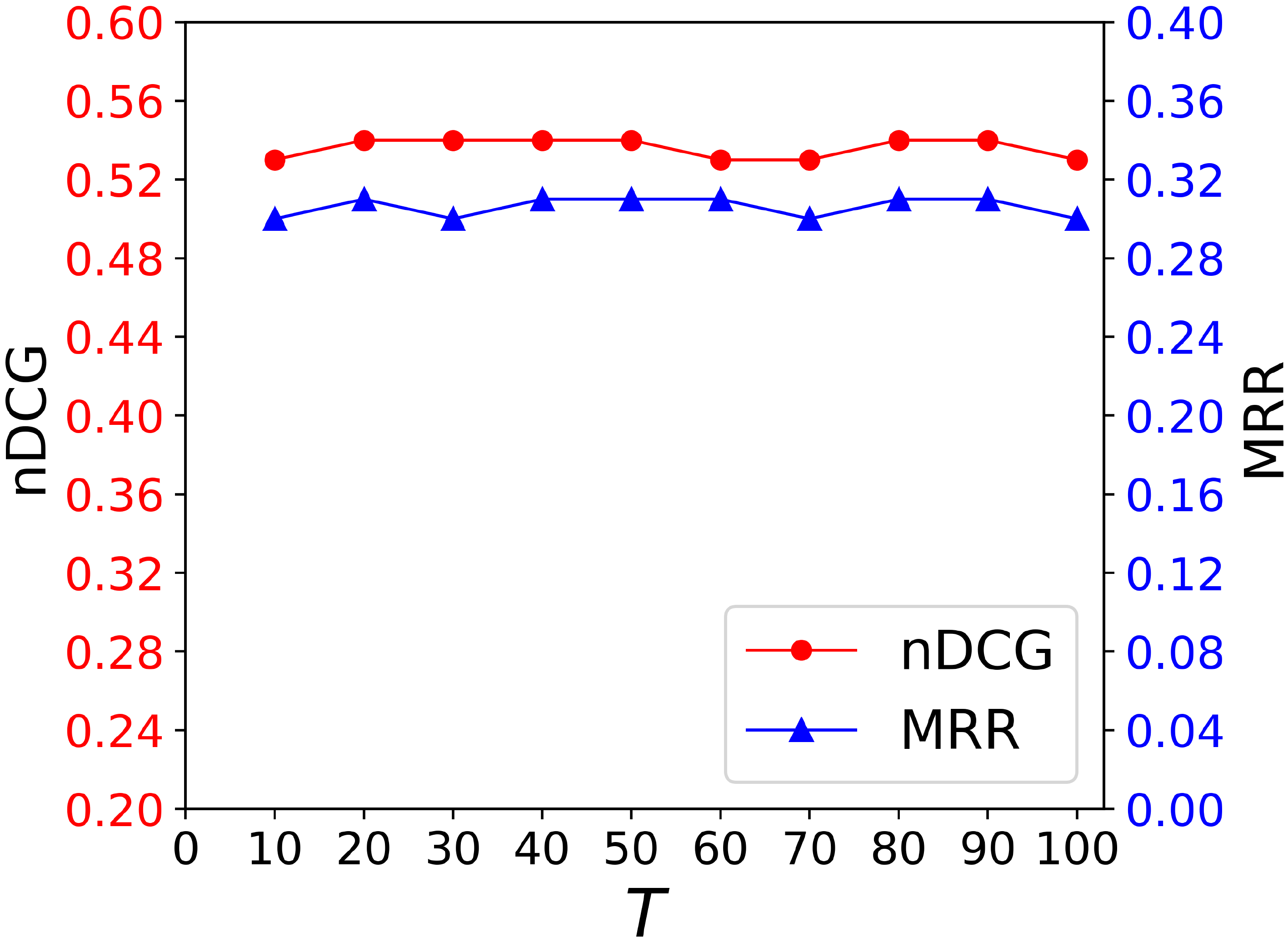}
			\caption{hobby, varying $T$.}
			\label{fig:d}
		\end{subfigure}
		\caption{Parameter study on personal attribute prediction from conversations.}
	\end{figure*}

	\subsection{Effectiveness Study} 

	We compare our proposed framework with the following personal attribute prediction methods. BERT IR \cite{dai2019deeper} trains BERT to calculate the relevance between an utterance $u_i$ and a document denoted by $d_{ext}$ from external data (e.g., Wiki-page or Wiki-category) w.r.t. an attribute value based on a binary cross-entropy loss. To fit the input size of BERT, $u_i$ and $d_{ext}$ are both split into 256-token chunks. CHARM \cite{tigunova2020charm} extracts keywords from $u_i$ by a cue detector and retrieves the relevant documents from external data by a value ranker. $\rm{CHARM_{BM25}}$ (resp. $\rm{CHARM_{KNRM}}$) adopts BM25 \cite{robertson1995okapi} (resp. KNRM \cite{xiong2017end}) as the value ranker of CHARM. No-keyword + BM25 uses $u_i$ and $d_{ext}$ directly as the input of BM25. RAKE (TextRank) + BM25 (KNRM) utilizes keywords extracted from $u_i$ by the unsupervised keyword extraction approach RAKE \cite{rose2010automatic} (TextRank \cite{mihalcea2004textrank}) and $d_{ext}$ as the input of BM25 (KNRM). HAM \cite{tigunova2019listening} can predict the score of the attribute value for $u_i$ by training the neural network based on the external data. ${\rm{HAM}_{avg}}$, $\rm{HAM_{2attn}}$, $\rm{HAM_{CNN}}$, and $\rm{HAM_{CNN-attn}}$ are four different configurations of HAM. $\rm{HAM_{CNN}}$/$\rm{HAM_{CNN-attn}}$ (resp. $\rm{HAM}_{avg}$/$\rm{HAM_{2attn}}$) adopts a text classification CNN (resp. two stacked fully connected layers). $\rm{HAM_{CNN-attn}}$/$\rm{HAM_{2attn}}$ (resp. $\rm{HAM}_{avg}$/$\rm{HAM_{CNN}}$) adopts attention mechanisms (resp. average methods) within and across utterances. DSCGN \cite{liu2022personal} fine-tunes BERT over unlabeled utterances and external data to predict an attribute value score list for $u_i$.
	
	Additionally, we add some recent SOTA weakly supervised text classification methods as baselines, which can be used to train an utterance classifier to predict the probability of attribute values for each utterance $u_i$. Specifically, ConWea \cite{mekala2020contextualized} can utilize user-provided seed words to create a contextualized utterance corpus, which is further leveraged to train an utterance classifier and expand seed words iteratively. SeedBTM \cite{yang2021dataless} could utilize user-provided seed words to extend BTM into an utterance classifier based on the word embedding technique. LOTClass \cite{meng2020text} generates some attribute-indicative words for each attribute value to fine-tune a PLM on a word-level category prediction task, and then does self-training on unlabeled utterances. X-Class \cite{wang2021x} can learn attribute-oriented utterance representations by a PLM and use the utterance-attribute value pairs generated by the Gaussian mixture model clustering process to train an utterance classifier. ClassKG \cite{zhang2021weakly} can generate pseudo labels for utterances by annotating keyword subgraphs, and train an utterance classifier with the pseudo labels.
	
	Apart from external data (e.g., Wiki-page or Wiki-category), the personal attribute prediction baselines BERT IR, $\rm{CHARM_{BM25}}$, and $\rm{CHARM_{KNRM}}$ require the labeled utterances to train the model and execute ten-fold cross-validation on each data set under a zero-shot setting in which the attribute values in the training set and the testing set are disjoint. Other personal attribute prediction baselines directly perform on the unlabeled utterances and external data without the requirement of labeled utterances, which is a relatively difficult setting, as it pushes ``zero-shot” to the extreme – no labeled utterances for any attribute values are provided. Compared with all the personal attribute prediction baselines, our framework PEARL and all the weakly supervised text classification methods perform on the unlabeled utterances only, which is an extremely difficult low-resource setting, as no other resources are provided except for the unlabeled utterances. Additionally, it is noted that some weakly supervised text classification methods require user-provided seed words, for each attribute value, we use its surface form as its seed word.
	
	From the results in Table \ref{effectiveness}, it can be seen that although consuming minimal resources, our proposed framework PEARL still yields the best performance compared with eighteen baselines on both data sets. To be specific, there are five weakly supervised text classification baselines with the same low-resource setting as us. Nevertheless, our method significantly surpasses them, which may be due to the fact that our framework can mine the personal attribute knowledge embedded in unlabeled utterances better. Despite that all the personal attribute prediction methods leverage more resources (e.g., labeled utterances or external data) than PEARL, PEARL still promotes by at least 2 (resp. 4) percentages compared with the best personal attribute prediction baseline DSCGN in terms of MRR (resp. nDCG) over both data sets. All the above experimental results validate the superiority of our proposed framework for predicting personal attributes from conversations. 
	\begin{table}[!t]
		\centering

		\scalebox{0.9}{\begin{tabular}{|c|cc|cc|}
				\hline
				\multirow{2}{*}{\textbf{\textit{Ablations}}} & \multicolumn{2}{c|}{\textbf{\textit{Profession}}} & \multicolumn{2}{c|}{\textbf{\textit{Hobby}}} \\
				\cline{2-5}          & \textbf{\textit{MRR}}   & \textbf{\textit{nDCG}}  & \textbf{\textit{MRR}}   & \textbf{\textit{nDCG}} \\
				\hline
				\hline
				PEARL & \textbf{0.49}  & \textbf{0.64}  & \textbf{0.31}  & \textbf{0.54} \\
				w/o AKI & 0.41  & 0.60   & 0.23  & 0.45 \\
				w/o the iteration operation & 0.45  & 0.62  & 0.29  & 0.52 \\
				\hline
		\end{tabular}}%
		\caption{Ablation study on personal attribute prediction from conversations.}
		\label{abla}%
	\end{table}%
	\subsection{Parameter Study}
	To investigate the robustness of our framework, we conduct sensitivity analysis to understand the impact of the parameter $K$ (i.e., the number of keywords for each utterance) and $T$ (i.e., the number of iterations for the Gibbs sampling process in Algorithm \ref{a1}) on our PEARL's performance over the profession and hobby data sets. From the trend plotted in Figure \ref{fig:a} and Figure \ref{fig:b}, we can see that when $K>50$ (resp. $K>60$), the performance achieved by PEARL is very stable on the profession (resp. hobby) data set, and is insensitive to the parameter $K$. From the trend plotted in Figure \ref{fig:c} and Figure \ref{fig:d}, it can be seen that the performance of PEARL is relatively stable on both data sets, and is insensitive to the parameter $T$.
	
	\begin{table}[!t]
	\centering
	
	\scalebox{1}{\begin{tabular}{|c|cc|}
			\hline
			\multirow{2}{*}{\textbf{\textit{Method}}} & \multicolumn{2}{c|}{\textbf{\textit{20News}}} \\
			\cline{2-3}          & \textbf{\textit{Micro-F1}} & \textbf{\textit{Macro-F1}} \\
			\hline
			\hline
			SeedBTM (2020) & 44.9  & 37.3 \\
			ConWea (2020) & 75.7 & 73.3 \\
			LOTClass (2020)& 73.8 & 72.5 \\
			X-Class (2021) & 78.6 & 77.8\\
			ClassKG (2021) & 83.8  & 82.7 \\
			\rowcolor{mygray}
			\Xhline{1pt}
			PEARL & \textbf{84.1} & \textbf{83.7} \\
			\rowcolor{mygray}
			w/o AKI  & 68.5  & 64.3 \\
			\rowcolor{mygray}
			w/o the iteration operation & 82.7  & 81.6\\
			\hline
	\end{tabular}}%
	\caption{Performance on the task of weakly supervised text classification. The performance of the baselines SeedBTM and ClassKG is reproduced via their open-source solutions. The results of the remaining baselines are taken from X-Class \cite{wang2021x}.}
	\label{tcp}%
\end{table}%
\subsection{Ablation Study}
To verify the importance of  different parts in our framework PEARL, we first remove the AKI module to make the BSA module work alone and perform attribute prediction by Formula \ref{pzd} directly (i.e., $P(b|d)$ is set to $1$). Considering that the AKI module alone cannot predict personal attributes from conversations, which has been discussed in the AKI part, so we cannot provide the effect of PEARL removing the BSA module. In addition, to examine the effectiveness of the iteration operation, we run PEARL by executing the BSA module first and thereafter executing the AKI module without the iteration operation (i.e., $E$ in Algorithm 1 is set to 1). 

From the results in Table \ref{abla}, we can draw the following observations: (1) without the AKI module, the performance of PEARL declines significantly on both data sets, which demonstrates the importance of the AKI module. This may be attributed to the fact that the biterm semantic information and the word co-occurrence information are complementary to some extent, and our framework PEARL can harness this complimentary knowledge effectively for better prediction; (2) without the iteration operation, PEARL performs worse over both data sets, which validates the point that the iteration operation can indeed obtain superior prior attribute knowledge derived from the AKI module to refine the Gibbs sampling process and enhance the prediction performance.

	\subsection{Experimental Analysis on Weakly Supervised Text Classification Task}
	To adapt our framework to the weakly supervised text classification task, we replace the attribute value (resp. utterance) with the class (resp. text) and select the class with the highest probability as the predicted label for the text based on the output ranking of PEARL. Following the previous weakly supervised text classification studies \cite{meng2020text,wang2021x}, we adopt the same evaluation metrics, i.e., micro-F1 and macro-F1, to evaluate PEARL, PEARL's two truncated versions, and five SOTA weakly supervised text classification methods over the common benchmark data set 20News \cite{lang1995newsweeder}, which consists of 17817 texts. Different from the profession (resp. hobby) data set containing 71 (resp. 149) attribute values as labels, 20News only has a fairly small number of five topics as labels (i.e., computer, sports, science, politics, and religion). 
	
	From the results in Table \ref{tcp}, it can be seen that (1) our framework PEARL exceeds all weakly supervised text classification methods, exhibiting its superiority and universality for the task of weakly supervised text classification; (2) PEARL outperforms its two truncated versions, which also demonstrates the effectiveness of the AKI module and the iteration operation for the text classification task.

\begin{figure}[!t]
		\centering
		\begin{subfigure}[t]{0.232\textwidth}
			\centering
			\includegraphics[width=\textwidth]{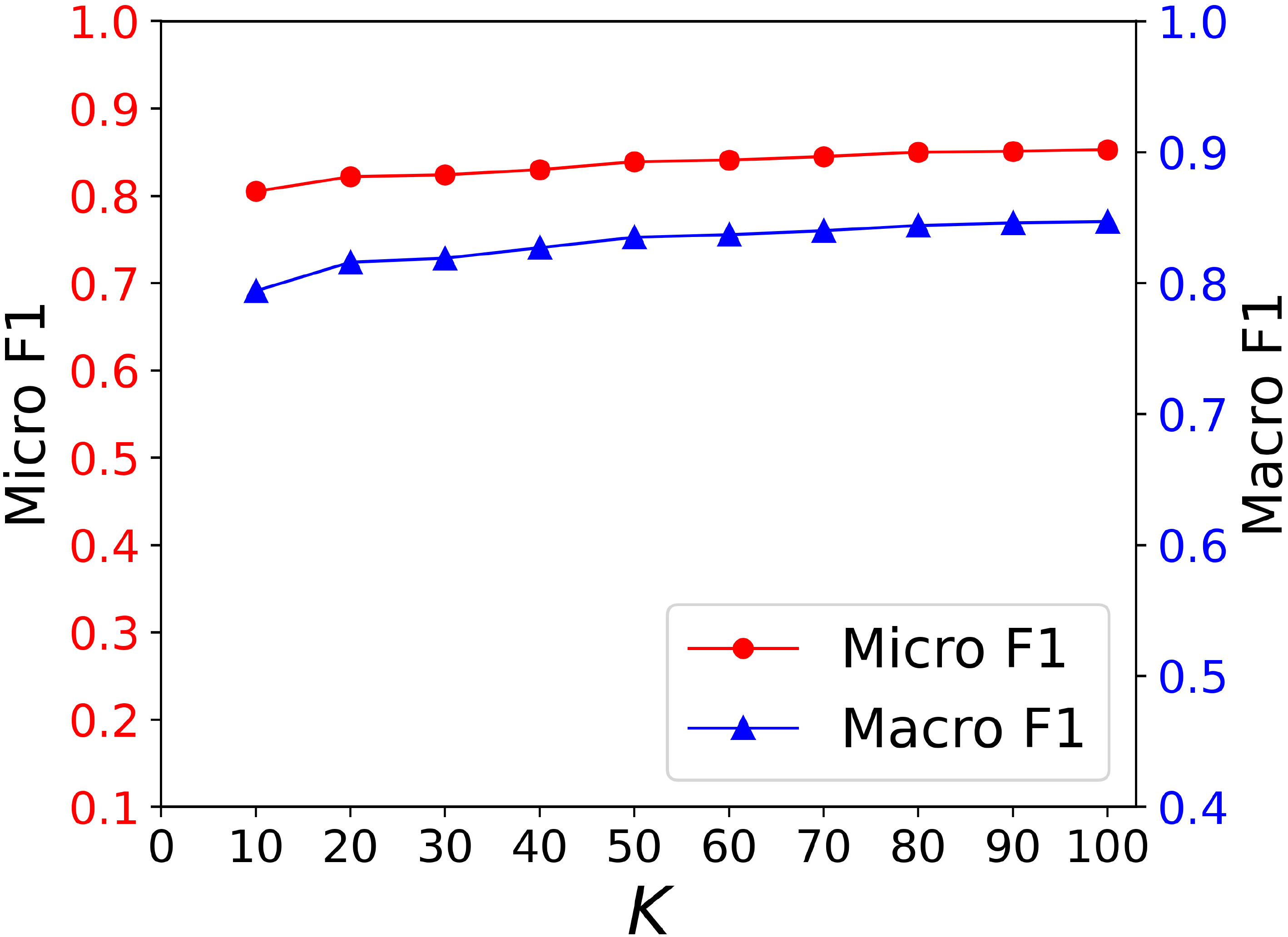}
			\caption{20News, varying $K$.}
			\label{sens1}
		\end{subfigure}
		\begin{subfigure}[t]{0.232\textwidth}
			\centering
			\includegraphics[width=\textwidth]{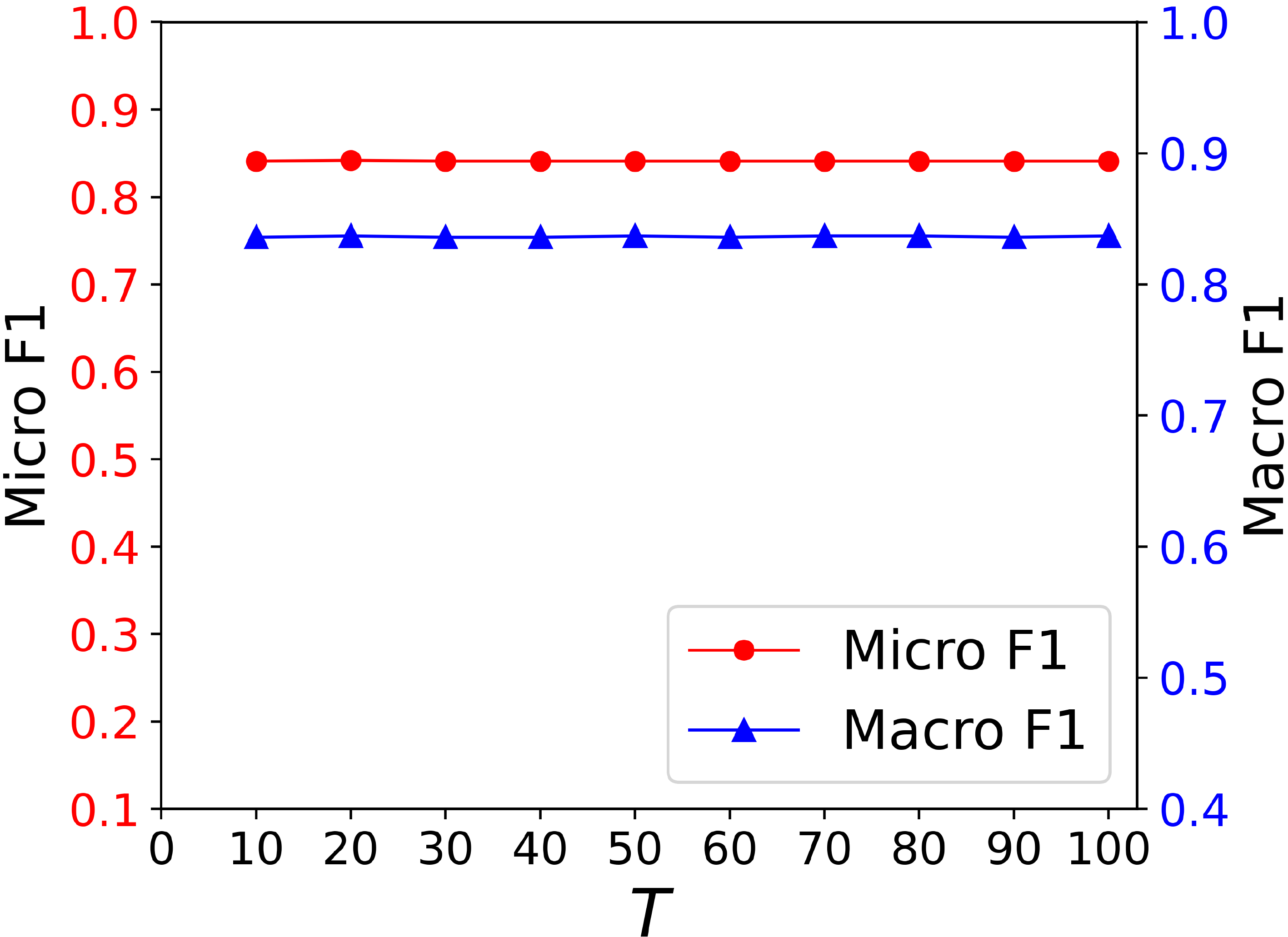}
			\caption{20News, varying $T$.}
			\label{sens2}
		\end{subfigure}
		\caption{Parameter study on weakly supervised text classification.}
	\end{figure}

	To explore the robustness of our framework on the weakly supervised text classification task, we further conduct sensitivity analysis to understand the impact of the parameters $K$ (i.e., the number of keywords for each utterance) and $T$ (i.e., the number of iterations for the Gibbs sampling process in Algorithm 1) on our PEARL's performance over the 20News data set. From the trend plotted in Figure \ref{sens1}, we can see that when $K>50$, the performance of PEARL changes little and is insensitive to the parameter $K$. From the trend plotted in Figure \ref{sens2}, it can be seen that PEARL performs relatively stable, and is insensitive to the parameter $T$.

	\section{Related Work}
\subsection{Personal Attribute Prediction}
The task of personal attribute prediction usually contains two aspects of studies: (1) personal attribute prediction from conversations; (2) personal attribute prediction from social media. We elaborate them as follows.
	
	Previous works extracted the profession attribute from conversations to build a PKB for a user by maximum-entropy classifiers \cite{jing2007extracting} and sequence-tagging CRFs \cite{li2014personal}. These methods assumed that users explicitly mentioned the attribute value in their utterances, so they are not applicable for the profession and hobby data sets without personal assertions. Recently, HAM \cite{tigunova2019listening} predicted scores of different attribute values for an utterance based on the stacked fully connected layers or CNNs by utilizing average approaches or attention mechanisms within and across utterances. CHARM \cite{tigunova2020charm} first extracted some keywords from utterances by leveraging BERT, and then matched these keywords against Web documents indicating possible attribute values from external data via some SOTA information retrieval ranking models (i.e., RAKE and KNRM). Specially, the procedure of the keyword extraction was trained by a reinforce policy gradient method. DSCGN \cite{liu2022personal} yielded two categories of supervision, i.e., document-level supervision via a distant supervision strategy and contextualized word-level supervision via a label guessing method from unlabeled utterances and external data, to fine-tune the language model with a noise-robust loss function. Different from all the above methods which consume plenty of resources, our framework does not rely on any labeled utterances or external data.
	
	Additionally, numerous works aim to predict personal attributes (e.g., age \cite{bayot2017age,mac2017demographic,liu2021age}, gender \cite{bayot2017age,mac2017demographic,vijayaraghavan2017twitter,basile2017n}, location \cite{shen2018predicting,liu2020named}, political preference \cite{vijayaraghavan2017twitter,preoctiuc2017beyond,xiao2020timme}, ethnicity \cite{preoctiuc2018user}, and occupational class \cite{preoctiuc2015analysis}) from social media such as Facebook and Twitter. The results provided by \cite{tigunova2019listening} show that three of these works \cite{basile2017n,bayot2017age,preoctiuc2015analysis} obtain unsatisfactory performance when predicting personal attributes from conversations, so they are not selected as our baselines. The remaining works rely on rich meta-data of social media (e.g., user profile, hashtag, and social network structure) that do not exist in conversation data, so they are unsuitable for our task.
	\subsection{Weakly Supervised Text Classification}
	In recent years, many methods have been proposed to use the label surface name as the weakly supervised signal to solve the text classification task without requiring any labeled documents or external data, whose setting is consistent with our proposed low-resource setting. Therefore, these methods can be utilized to predict personal attributes by training an utterance classifier based on unlabeled utterances. These weakly supervised text classification approaches \cite{meng2020text, mekala2020contextualized,wang2021x,zhang2021weakly} usually trained a classifier by using the pseudo labels and refined the model over the unlabeled data via self-training. However, they perform not well on the profession and hobby data sets, which has been verified in our experiments. 
	

	\section{Conclusion and Future Work}
	To predict personal attributes from conversations under a low-resource setting which does not resort to any labeled utterances or external data, we propose a novel framework PEARL that combines the biterm semantic information with the word co-occurrence information seamlessly in an iterative manner. Extensive experiments have demonstrated the effectiveness of our framework PEARL against many SOTA personal attribute prediction methods and weakly supervised text classification methods. In addition, we argue that the profession and hobby data sets can be utilized to measure the efficacy of weakly supervised text classification methods to a certain extent, which can benefit the text classification research community for future studies.
\section{Acknowledgments}
This work was supported in part by National Natural Science Foundation of China (No. U1936206, 62272247), YESS by CAST (No. 2019QNRC001), and CAAI-Huawei MindSpore Open Fund. 
\bibliography{aaai23}

\end{document}